\documentclass[twocolumn]{cinc}
\usepackage{graphicx}

\usepackage{url}

\usepackage{breakurl}
\usepackage[breaklinks]{hyperref}

\begin{document}
\bibliographystyle{cinc}

\title{A Comparative Study of Diabetes Prediction Based on Lifestyle Factors \\Using Machine Learning}


\author {Bruce Nguyen, Yan Zhang \\
\ \\ 
School of Computer Science and Engineering\\ 
California State University San Bernardino\\
007588120@coyote.csusb.edu, Yan.Zhang@csusb.edu}

\maketitle

\begin{abstract}

Diabetes is a prevalent chronic disease with significant health and economic burdens worldwide. Early prediction and diagnosis can aid in effective management and prevention of complications. This study explores the use of machine learning models to predict diabetes based on lifestyle factors using data from the Behavioral Risk Factor Surveillance System (BRFSS) 2015 survey. The dataset consists of 21 lifestyle and health-related features, capturing aspects such as physical activity, diet, mental health, and socioeconomic status. Three classification models — Decision Tree, K-Nearest Neighbors (KNN), and Logistic Regression — are implemented and evaluated to determine their predictive performance. The models are trained and tested using a balanced dataset, and their performances are assessed based on accuracy, precision, recall, and F1-score. The results indicate that the Decision Tree, KNN, 
and Logistic Regression achieve an accuracy of 74\%, 72\%, and 75\%, respectively, with varying strengths in precision and recall. 
The findings highlight the potential of machine learning in diabetes prediction and suggest future improvements through feature selection and ensemble learning techniques.
    
\end{abstract}

\section{Introduction}

Diabetes is a chronic disease that affects millions of people around the world, with significant health and economic implications~\cite{roglic2016global}. Early detection and intervention can help manage the disease and prevent complications. Traditional diagnostic methods are based on clinical tests and medical history, but recent advances in machine learning offer promising approaches to predict diabetes based on lifestyle factors~\cite{chaki2022machine}.

This study aims to compare different machine learning models for diabetes prediction using data from the Behavioral Risk Factor Surveillance System (BRFSS) 2015 survey~\cite{cdc_ds}. The dataset includes various lifestyle and health-related attributes such as high blood pressure, cholesterol levels, physical activity, diet, and socioeconomic status. Three classification models are evaluated for their effectiveness in predicting diabetes and prediabetes: Decision Tree, K-Nearest Neighbors (KNN), and Logistic Regression.

The paper is structured as follows: Section 2 reviews related work on machine learning applications in diabetes prediction. Section 3 describes the dataset and preprocessing steps. Section 4 details the methodologies employed for classification. Section 5 presents the experimental results and model evaluations. Finally, Section 6 concludes the study and discusses future research directions.

\section{Related Work}
As the diabetes pandemic has become more widespread in the last few decades, so have the efforts in the field. 
Many companies and researchers invest part of their resources in trying to predict diabetes. 
Machine learning techniques have increasingly been applied to the prediction of diabetes, offering improvements in accuracy, efficiency, and automation compared to traditional diagnostic methods. 

Mujumdar and Vaidehi highlighted the role of big data analytics in healthcare and proposed a predictive model incorporating both traditional and external factors such as glucose levels, BMI, insulin, and age~\cite{mujumdar2019diabetes}. The study emphasized the need for improved classification accuracy over existing methods and introduces a pipeline model to improve predictive performance. 
The results demonstrated that
the inclusion of additional features contributes to better classification outcomes~\cite{mujumdar2019diabetes}.

Chaki et al. provided an extensive review of more than 100 publications on the use of machine learning and artificial intelligence for the detection, diagnosis and self-management of diabetes~\cite{chaki2022machine}. 
The review categorized methodologies based on dataset selection, preprocessing techniques, feature extraction, and model performance measures. It also identified research gaps in personalization and self-management tools, indicating the need for developments in patient-specific predictive models~\cite{chaki2022machine}.

Kopitar, et al. compared machine learning-based prediction models, including Glmnet, Random Forest (RF), XGBoost, and LightGBM, with traditional regression-based screening methods~\cite{kopitar2020early}. The study found that while machine learning models showed promise, they did not provide clinically significant improvements over simpler regression models when predicting fasting plasma glucose levels. 
The study underscored the importance of balancing model interpretability with predictive performance in clinical applications~\cite{kopitar2020early}.

Lai, et al. focused on building predictive models for Canadian patients using Logistic Regression and Gradient Boosting Machine (GBM)~\cite{lai2019predictive}. 
The study evaluated the models using AROC scores, demonstrating that GBM (84.7\%) and Logistic Regression (84.0\%) outperform Decision Tree and Random Forest models. 
The study highlighted the significance of fasting blood glucose, BMI, high-density lipoprotein, and triglycerides as the most important predictors~\cite{lai2019predictive}. 

Khanam and Foo investigated various machine learning algorithms using the Pima Indian Diabetes (PID) dataset~\cite{khanam2021comparison}. 
The study evaluated seven different machine learning models and identified Logistic Regression and Support Vector Machine as the most effective algorithms for diabetes prediction. 
The study developed a Neural Network  model with varying hidden layers, finding that the model with 2 hidden layers achieves 88.6\% accuracy~\cite{khanam2021comparison}. 

Parimala, et al. examined numerous machine learning algorithms in an attempt to predict diabetes~\cite{parimala2023diabetes}. 
The study revolved around building classification models for the diabetes dataset acquired, using these results to develop better models to determine whether or not a person has diabetes. The results revealed the Random Forest machine as the most efficient model with an accuracy of 98\%~\cite{parimala2023diabetes}.

Tasin, et al. analyzed several machine learning to predict diabetes based on the Pima Indian Diabetes (PID) dataset~\cite{tasin2023diabetes}.
SMOTE and ADASYN approaches were employed to manage the class imbalance problem.
The study found that XGBoost classifier with ADASYN obtained 81\% accuracy, Bagging classifier with SMOTE approach following close behind with 79\% accuracy~\cite{tasin2023diabetes}.

In summary, these studies demonstrate the potential of machine learning for diabetes prediction, with each offering unique insights into model selection, feature importance, and performance evaluation. 
Future research should focus on improving model robustness, incorporating real-time patient data, and enhancing personalization to optimize diabetes prediction and management.

\section{Datasets and Preprocessing}
The study uses data from the Behavioral Risk Factor Surveillance System (BRFSS) 2015 survey conducted by the Centers for Disease Control and Prevention (CDC)~\cite{cdc_ds}. 
The original dataset comprises 441,455 survey responses with 330 features, which include both direct participant responses and calculated variables. After cleaning, the dataset was reduced to 253,680 responses with 21 feature variables and a target variable classifying respondents into three categories: no diabetes (class 0), prediabetes (class 1), and diabetes (class 2)~\cite{kaggle_ds}.
However, this cleaned dataset are  unbalanced, with the majority of respondents falling into the "No diabetes" category. 

To address this issue, a balanced version of the dataset was created, containing 70,692 survey responses with an equal 50-50 split of respondents with no diabetes (class 0) and those with either prediabetes or diabetes (class 1)~\cite{kaggle_ds}. 
The balanced dataset retains the 21 feature variables, which capture a wide range of lifestyle and health-related factors, such as high blood pressure, cholesterol levels, physical activity, diet, mental and physical health, and socioeconomic status.
From the 21 feature variables, 14 are binary types and the remaining 7 are integer types. 
A brief overview of the dataset can be seen in Table~\ref{tab:feature_description}.
\begin{table*}[htbp]
\centering
\caption{Feature Description}
\label{tab:feature_description}
\begin{tabular}{|l|l|l|p{8cm}|}
\hline
\textbf{Attribute Name} & \textbf{Data Type} & \textbf{Values} & \textbf{Description} \\ \hline
Diabetes\_binary & Binary & 0, 1 & 0: No Diabetes, 1: Prediabetes/Diabetic \\ \hline
HighBP & Binary & 0, 1 & Presence of High Blood Pressure  \\ \hline
HighChol & Binary & 0, 1 & Presence of High Cholesterol \\ \hline
CholCheck & Binary & 0, 1 & Cholesterol check in the last 5 years \\ \hline
BMI & Integer & 12-98 & Body Mass Index (BMI) \\ \hline
Smoker & Binary & 0, 1 & Smoked 100+ cigarettes in entire life \\ \hline
Stroke & Binary & 0, 1 & Ever had a stroke? \\ \hline
HeartDiseaseorAttack & Binary & 0, 1 & Coronary Heart Disease or Myocardial Infarction \\ \hline
PhysActivity & Binary & 0, 1 & Physical activity in past 30 days \\ \hline
Fruits & Binary & 0, 1 & Consumed fruits 1+ times per day \\ \hline
Veggies & Binary & 0, 1 & Consumed vegetables 1+ times per day \\ \hline
HvyAlcoholConsumption & Binary & 0, 1 & Men $\geq$ 14 drinks per week, Women $\geq$ 7 drinks per week \\ \hline
AnyHealthcare & Binary & 0, 1 & Health care coverage (including health insurance, etc.) \\ \hline
NoDocbcCost & Binary & 0, 1 & Could not visit doctor in past 12 months because of cost \\ \hline
GenHlth & Integer & 1-5 & Rate general health on a scale 1-5: \{1 = Excel., 5 = Poor\} \\ \hline
MentHlth & Integer & 0-30 & Days of poor mental health \{1-30 days\} \\ \hline
PhysHealth & Integer & 0-30 & Days of physical illness or injury \{1-30 days\} \\ \hline
DiffWalk & Binary & 0, 1 & Difficulty walking/climbing stairs  \\ \hline
Sex & Binary & 0, 1 & Individual’s Sex \{0: Female, 1: Male\} \\ \hline
Age & Integer & 1-13 & Individual’s Age broken up into 13 categories \\ \hline
Education & Integer & 1-6 & Individual’s Education level broken into 6 categories \\ \hline
Income & Integer & 1-8 & Individual’s Income level broken into 8 categories \\ \hline
\end{tabular}
\end{table*}

Some machine learning methods, such as K-Nearest Neighbors (KNN) and Logistic Regression, are sensitive to the scale of the input data. Therefore, numerical features like BMI, GenHlth, MentHlth,  PhysHealth, Age, Education, and Income are standardized to lie between 0 and 1 to ensure the dataset's suitability for all classification models.

\section{Methodologies}
The three machine learning algorithms implemented are Decision Trees, K-Nearest Neighbors (KNN), and Logistic Regression. 
Each algorithm has its own strengths and limitations, with different parameters that can be tuned to optimize performance. 
These algorithms were chosen for their interpretability and scalability for the diabetes dataset.

\subsection{Decision Tree Classification}
Decision Trees are a supervised learning algorithm used for classification and regression tasks. 
They work by recursively splitting the dataset into subsets based on the most significant attribute at each node, as determined by impurity metrics such as Gini or Entropy~\cite{shalev2014understanding}. This process continues until a stopping criterion is met, such as reaching a maximum tree depth or maximum leaf nodes. 
Decision Trees are particularly advantageous for datasets with non-linear relationships, as they can capture complex interactions between features without requiring extensive preprocessing~\cite{shalev2014understanding}. 
However, they are prone to overfitting, especially with deep trees, which can be avoid through tuning parameter values. 
In this study, the Decision Trees are trained with the hyperparameter maximum leaf nodes tuned to optimize performance on the diabetes dataset.

\subsection{K-Nearest Neighbors (KNN)}
K-Nearest Neighbors (KNN) is an instance-based, non-parametric learning algorithm that classifies a data point based on the majority class or average value of its "k" nearest neighbors in the feature space~\cite{shalev2014understanding}. The distance metric, typically Euclidean distance, is used to identify the nearest neighbors. 
KNN is effective for datasets with clear clustering patterns and is well-suited for smaller datasets due to its computational simplicity. 
KNN is sensitive to the scale of the input features. 
To address this, numerical features in the diabetes dataset are normalized using min-max scaling before applying KNN. 
The choice of "k" (the number of neighbors) is critical; a small "k" may lead to overfitting, while a large "k" may oversmooth the decision boundaries. In this study, cross-validation is used to determine the optimal value of "k" for the diabetes dataset.

\subsection{Logistic Regression}
Logistic Regression is a parametric algorithm used for binary and multi-class classification problems. 
It models the relationship between the dependent variable and independent variables using the Sigmoid Function, which outputs probabilities between 0 and 1~\cite{pampel2020logistic}. 
These probabilities are then thresholded to make class predictions. 
Logistic Regression is effective for datasets with linear relationships between features and the target class. It is computationally efficient and provides interpretable results through the coefficients of the input features, which indicate their relative importance~\cite{pampel2020logistic}. However, it assumes linearity between the features and target variable, which may limit its performance on datasets with non-linear relationships. 
To enhance its performance, regularization techniques such as L1 (Lasso) or L2 (Ridge) regularization can be applied to prevent overfitting. 
In this study, Logistic Regression is implemented with hyperparameter regularization strength tuned to achieve optimal classification performance.

\section{Experimental Results and Discussion}
We split the dataset into training and testing sets, with 80\% of the data (around 56,500 instances) used for training and 20\% (approximately 14,100 instances) reserved for testing. 
This split ensured that the models were trained on a substantial portion of the data while leaving enough for performance evaluation on unseen data.
Scikit-learn machine learning library is used to train 3 machine learning models~\cite{pedregosa2011scikit}.

\subsection{Decision Tree Training and Evaluation}
\begin{figure}[htbp]
    \centering
    \includegraphics[width=8cm]{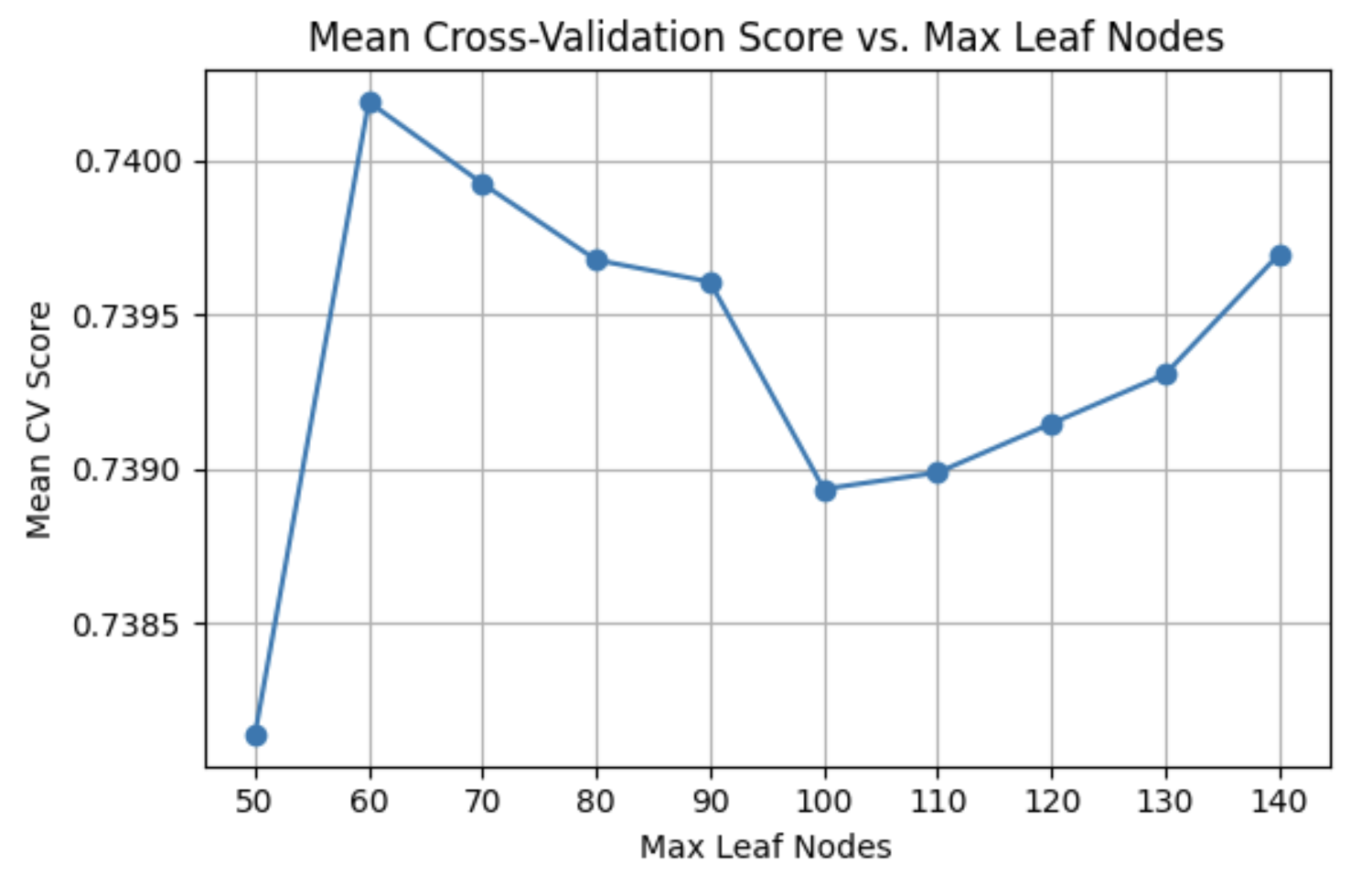}
    \caption{Decision Tree Cross Validation}
    \label{figure:DT_CV}
\end{figure}

When training decision trees, Gini was used to measure the impurity of nodes.
To optimize the performance of the Decision Tree model, the hyperparameter maximum leaf nodes was tuned using grid search with cross-validation. 
The best-performing model was found to have 60 leaf nodes, achieving the highest cross-validation score, as shown in Figure~\ref{figure:DT_CV}.

The trained Decision Tree model was tested on the test dataset. 
The confusion matrix for the test results is shown in Table~\ref{tab:dt_cm}.
\begin{table}[htbp]
    \centering
    \caption{The confusion matrix of Decision Tree.}
    \begin{tabular}{c|cc}
\hline\textbf{True/Predicted}  & \textbf{0} & \textbf{1} \\
        \hline
        \textbf{0} & 5060 & 1955  \\
        \textbf{1} & 1719   & 5405 \\
        \hline
    \end{tabular}
    \label{tab:dt_cm}
\end{table}
The classification report summarizing the model's performance is provided in Table~\ref{tab:dt_cr}.
\begin{table}[htbp]
    \centering
    \caption{The classification report of Decision Tree.}
    \begin{tabular}{c|cccc}
        \hline
        \textbf{Class} & \textbf{Precision} & \textbf{Recall} & \textbf{F1} & \textbf{Support} \\
        \hline
        \textbf{0} & 0.75 & 0.72 & 0.73 & 7015 \\
        \textbf{1} & 0.73 & 0.76 & 0.75 & 7124 \\
        \hline
        \textbf{Accuracy} & \multicolumn{3}{c}{0.74} & 14139 \\
        \hline
    \end{tabular}
    \label{tab:dt_cr}
\end{table}

The Decision Tree model achieved an accuracy of 74\%, with balanced precision and recall values across both classes. The F1-score of 0.75 for class 1 (prediabetes or diabetes) indicates the model's ability to detect individuals at risk while maintaining a reasonable trade-off between precision and recall. 
The confusion matrix suggests that the model correctly classified most cases, though some misclassification still occurred, particularly in distinguishing between class 0 and class 1. Further tuning or incorporating additional features may enhance the model's predictive performance.

\subsection{KNN Training and Evaluation}

\begin{figure}[htbp]
    \centering
\includegraphics[width=8cm]{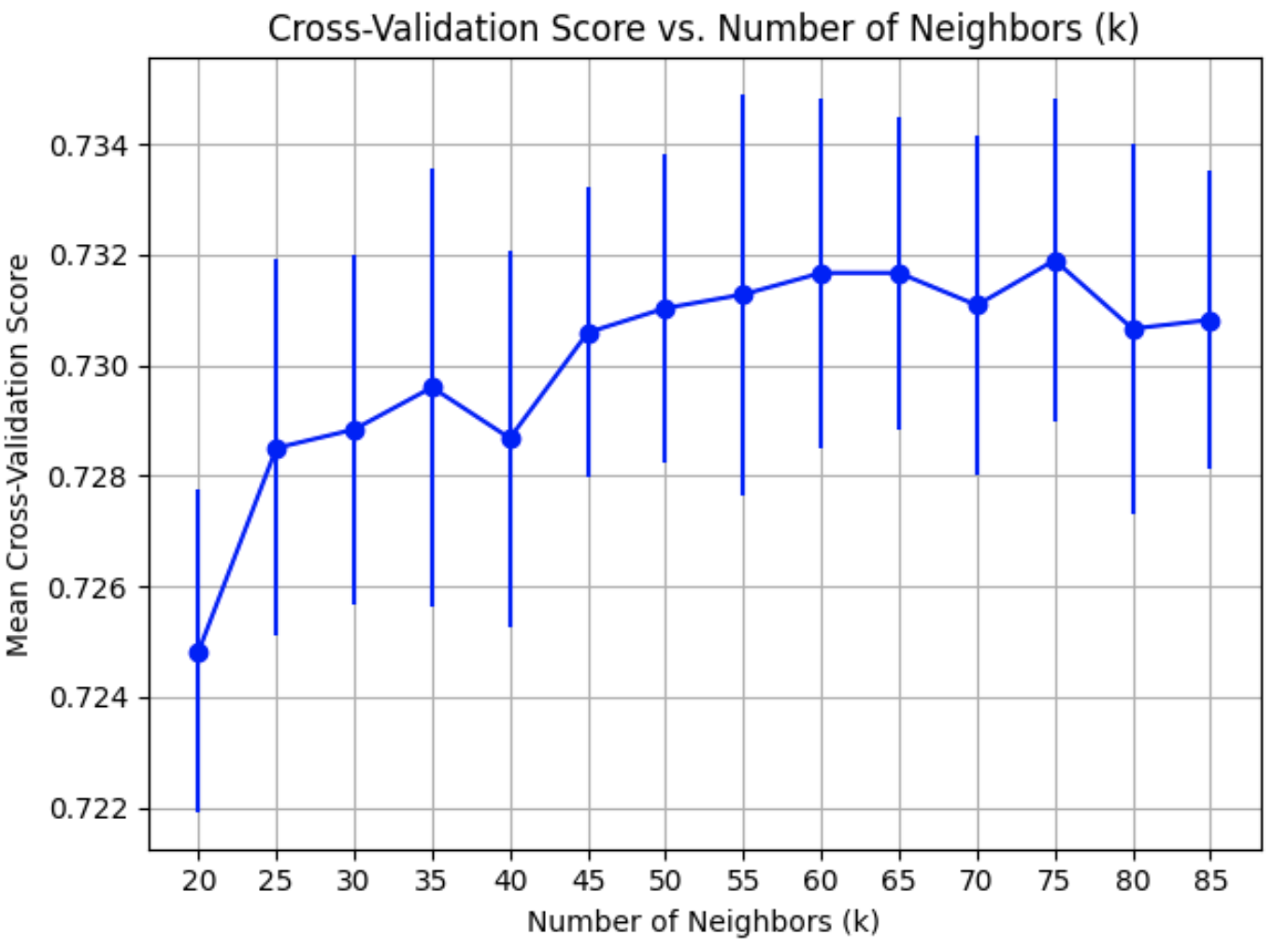}
    \caption{KNN Cross Validation}
    \label{figure:KNN_CV}
\end{figure}

For the K-Nearest Neighbors (KNN) model, Euclidean distance was used to determine the nearest neighbors, and all points within each neighborhood were weighted equally. 
The optimal k value was selected using 10-fold cross-validation, with $k = 75$ achieving the highest cross-validation score, as shown in Figure~\ref{figure:KNN_CV}.

The trained KNN model was then evaluated on the test dataset. The confusion matrix for the test results is presented in Table~\ref{tab:knn_cm}.
\begin{table}[htbp]
    \centering
    \caption{The confusion matrix of KNN.}
    \begin{tabular}{c|cc}
\hline\textbf{True/Predicted}  & \textbf{0} & \textbf{1} \\
        \hline
        \textbf{0} & 4765 & 2250  \\
        \textbf{1} & 1642   & 5482 \\
        \hline
    \end{tabular}
    \label{tab:knn_cm}
\end{table}
The classification report for the KNN model is shown in Table~\ref{tab:knn_cr}.
\begin{table}[htbp]
    \centering
    \caption{The classification report of KNN.}
    \begin{tabular}{c|cccc}
        \hline
        \textbf{Class} & \textbf{Precision} & \textbf{Recall} & \textbf{F1} & \textbf{Support} \\
        \hline
        \textbf{0} & 0.74 & 0.68 & 0.71 & 7015 \\
        \textbf{1} & 0.71 & 0.74 & 0.74 & 7124 \\
        \textbf{Accuracy} & \multicolumn{3}{c}{0.72} & 14139 \\
        \hline
    \end{tabular}
    \label{tab:knn_cr}
\end{table}

The KNN model achieved an accuracy of 72\%. The F1-score for class 1 (prediabetes or diabetes) is 0.74, indicating the model's effectiveness in correctly identifying at-risk individuals. However, compared to the Decision Tree model, the KNN model has a lower recall for class 0, suggesting that it misclassifies more non-diabetic cases. The performance may potentially be improved by including more features or more instances to enhance model performance.

\subsection{Logistic Regression Training and Evaluation}

For training the Logistic Regression model, the "lbfgs" solver was used, which is well-suited for high-dimensional data. L2 regularization was applied to manage sparsity, reduce the risk of overfitting by penalizing excessively large coefficients in the model. The model was trained for up to 100 iterations to allow the "lbfgs" solver to converge to a stable solution.

The trained Logistic Regression model was then evaluated on the test dataset. The confusion matrix for the test results is presented in Table~\ref{tab:lg_cm}.
\begin{table}[htbp]
    \centering
    \caption{The confusion matrix of Logistic Regression.}
    \begin{tabular}{c|cc}
\hline\textbf{True/Predicted}  & \textbf{0} & \textbf{1} \\
        \hline
        \textbf{0} & 5125 & 1890  \\
        \textbf{1} & 1679   & 5445 \\
        \hline
    \end{tabular}
    \label{tab:lg_cm}
\end{table}
The classification report for the Logistic Regression model is shown in Table~\ref{tab:lg_cr}.
\begin{table}[htbp]
    \centering
    \caption{The classification report of Logistic Regression.}
    \begin{tabular}{c|cccc}
        \hline
        \textbf{Class} & \textbf{Precision} & \textbf{Recall} & \textbf{F1} & \textbf{Support} \\
        \hline
        \textbf{0} & 0.75 & 0.73 & 0.74 & 7015 \\
        \textbf{1} & 0.74 & 0.76 & 0.75 & 7124 \\
        \hline
        \textbf{Accuracy} & \multicolumn{3}{c}{0.75} & 14139 \\
        \hline
    \end{tabular}
    \label{tab:lg_cr}
\end{table}

The Logistic Regression model achieved an accuracy of 75\%, slightly outperforming the Decision Tree and KNN models. 
The model demonstrated balanced precision and recall values, indicating its robustness in distinguishing between diabetic and non-diabetic cases.

\subsection{Model Comparison and Discussion} 
Comparing the three models, Logistic Regression achieved the highest accuracy of 75\%, outperforming Decision Tree (74\%) and KNN (72\%). 
The Logistic Regression model demonstrated balanced precision and recall values.

While all three models performed similarly in terms of overall predictive ability, the Decision Tree and Logistic Regression had slightly higher recall for class 1, making them more sensitive in identifying at-risk individuals. 
The KNN model showed the lowest recall for class 0, indicating a higher tendency to misclassify non-diabetic cases. 
The choice of model depends on the application needs — whether prioritizing recall for high-risk cases or maintaining a balance between precision and recall.

\section{Conclusion}
This study compared Decision Tree, K-Nearest Neighbors (KNN), and Logistic Regression models for diabetes prediction using the BRFSS 2015 dataset. The results showed that Logistic Regression achieved the highest accuracy (75\%), followed by Decision Tree (74\%) and KNN (72\%). While all models demonstrated reasonable predictive capability, Logistic Regression provided the most balanced performance in terms of precision and recall.

Future work could explore the inclusion of additional features, such as more detailed dietary habits, to enhance model performance. 
Additionally, employing ensemble learning techniques or deep learning models may further improve classification performance.

\balance

\bibliography{main}


  
  
      

\end{document}